\documentclass{article}
\usepackage{spconf,amsmath,graphicx,hyperref}
\usepackage{booktabs}
\usepackage{comment}
\usepackage[ruled,vlined]{algorithm2e} 

\title{KCM: KAN-based collaboration models enhance pretrained large models}
\name{Guangyu Dai, Siliang Tang, Yueting Zhuang}
\address{Zhejiang University}
\begin{document}
\ninept
\maketitle
\begin{abstract}
In recent years, Pretrained Large Models(PLMs) have demonstrated exceptional performance across diverse tasks in multiple domains. Nevertheless, these large-scale models face challenges including substantial computational overhead, API dependency, and inaccuracies in domain-specific knowledge. Consequently, researchers proposed large-small model collaboration frameworks, leveraged easily trainable small models to assist large models, aiming to (1) reduce computational resource consumption while maintaining comparable accuracy, and (2) enhance large model performance in specialized domain tasks. However, this collaborative paradigm suffers from issues such as significant accuracy degradation, exacerbated catastrophic forgetting, and amplified hallucination problems induced by small model knowledge. To address these challenges, we propose a KAN-based Collaborative Model (KCM) as an improved approach to large-small model collaboration. The KAN utilized in KCM represents an alternative neural network architecture distinct from conventional MLPs. Compared to MLPs, KANs exhibit superior visualizability and interpretability while mitigating catastrophic forgetting. We deployed KCM in large-small model collaborative systems across three scenarios: language, vision, and vision-language cross-modal tasks. The experimental results demonstrate that, compared with pure large model approaches, the large-small model collaboration framework utilizing KCM as the collaborative model significantly reduces the number of large model inference calls while maintaining near-identical task accuracy, thereby substantially lowering computational resource consumption. Concurrently, the KAN-based small collaborative model markedly mitigates catastrophic forgetting, leading to significant accuracy improvements for long-tail data. Furthermore, in ablation studies, we compared the large-small model collaboration methods employing MLP-based small collaborative models (MCM) against those using KAN-based collaborative models (KCM). The results reveal that KCM demonstrates superior performance across all metrics compared to MCM.

\end{abstract}
\begin{keywords}
KAN, collaborative model, data distillation
\end{keywords}
\section{Introduction}
\label{sec:intro}
Recently, Pre-trained Large Models (PLMs) have revolutionized artificial intelligence research. These models, trained on vast datasets, adapt to numerous downstream tasks—such as GPT in language modality and BLIP-2 in vision-language modality. With their widespread adoption, inherent limitations have emerged, including high computational costs and insufficient domain-specific knowledge. Consequently, researchers have proposed collaborative frameworks where small models assist large models to mitigate these issues. Examples include leveraging small models for Retrieval-Augmented Generation (RAG) tasks\cite{asai2024selfrag,tan-etal-2024-small,yan2024corrective}, facilitating In-context Learning \cite{xu-etal-2024-small}, and reducing overall computational expenditure through MLP-based collaborative small models\cite{jiang2024longllmlingua,chen2024data}. Studies reveal that while small models underperform large models on identical tasks, this performance gap primarily manifests in challenging samples (e.g. tail region samples in long-tail datasets). Thus, classifying data and substituting small models for large models on specific subsets significantly reduces computational costs.\\
This study introduces KAN-based Collaborative Models (KCMs), which reduce overall computational costs by minimizing large model invocations without compromising accuracy. Specifically, KCMs employ a small model to discriminate input data: relatively simple samples are processed by the small model, while complex samples are routed to the large model. Additionally, we propose bidirectional synergy between small and large models to enhance their respective performance. At the small model part, when samples are transferred to the large model, the reduced likelihood of belonging to the small model’s proficient distribution is incorporated into prompts to augment the large model’s decision-making. Conversely, when samples transition from the large model to the small model, knowledge distillation improves the small model’s efficacy. To counter catastrophic forgetting, KAN—known for its robustness against this issue—is adopted as the foundational architecture for our small model. Experiments were conducted across three scenarios: language modality, visual modality, and vision-language multimodality. In ablation studies, KCMs demonstrated significant computational cost reduction compared to MLP-based Collaborative Models (MCMs), alongside slight accuracy improvements.\\
The primary contributions of this paper are as follows:
\begin{itemize}
    \item KCM+PLM paradigm achieves comparable or minimally acceptable accuracy degradation relative to pure large model approaches, while consuming substantially fewer computational resources;
    \item KCM exhibits remarkable efficacy in alleviating catastrophic forgetting for long-tailed tasks; 
    \item KCM outperforms MLP-based collaborative components when serving as small collaborative model.
\end{itemize}
\section{Related works}
\label{sec:re-works}
\subsection{Large-Small Model Collaboration Methods}
Large Language Models (LLMs) have exhibited remarkable proficiency across diverse NLP tasks in recent years, including text generation, question answering, and reasoning. Concurrently, Vision-Language Models (VLMs) have delivered unprecedented performance in vision-text multimodal tasks within the vision and multimodal domains. However, these large-scale models demand massive parameter scales and substantial computational resources, while demonstrating limited efficiency in domain-specific applications. This challenge has motivated researchers to leverage small models as auxiliary components for large models, aiming to enhance scenario-specific performance by integrating outputs from computationally efficient and easily trainable small models with those of large models. Such approaches—wherein small models assist large models—have emerged in domains \cite{wang2024comprehensive}, including Retrieval-Augmented Generation (RAG) \cite{asai2024selfrag,huang-etal-2024-less}, In-Context Learning (ICL) \cite{xu-etal-2024-small}, language translation and summarization tasks \cite{bergner2024think}.
\subsection{KAN Introduction}
Kolmogorov-Arnold Networks(KANs)\cite{liu2024kan} is a novel neural network architecture grounded in the Kolmogorov-Arnold representation theorem, fundamentally distinct from traditional Multi-Layer Perceptrons (MLPs). The difference between MLPs and KANs can be seen in fig.1. The defining characteristic of KANs is the placement of learnable activation functions on the network’s edges, while summation operations occur at the nodes. In contrast, MLPs employ fixed activation functions at the nodes, with edges performing solely linear transformations. Algorithmically, this shift renders the activation functions in KANs learnable, unlike their fixed counterparts in MLPs. This learnability endows KANs with superior visualizability, interpretability, and interactivity. Furthermore, KANs can potentially mitigate the issue of catastrophic forgetting, offering a significant advantage over MLPs in several domains. However, these benefits come at the cost of increased computational expenses and reduced training efficiency.\\
\begin{figure}[!h]
  \centering
  \includegraphics[scale=0.3]{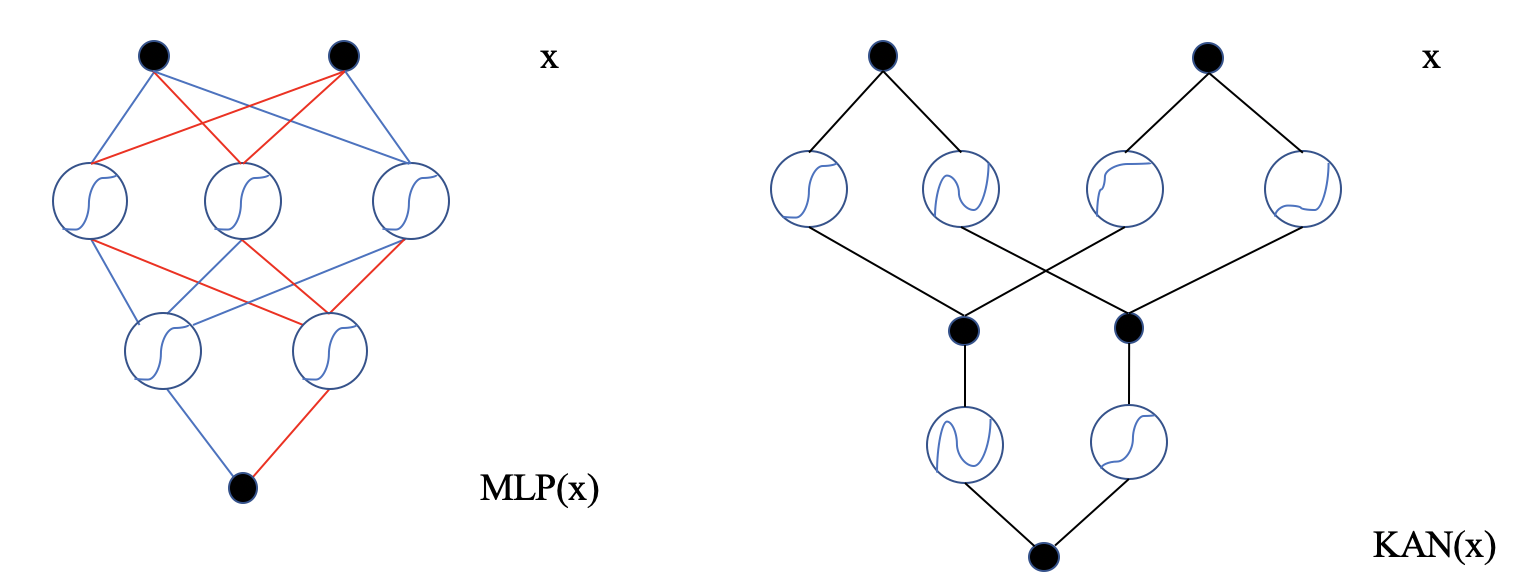}
  \caption{The overview of MLP and KAN.MLP performs polynomial calculation at the node, and the activation function is the same; KAN is calculated on the edge, the activation functions may be different, and the nodes of KAN only play the role of addition.}
  \label{fig:fig1}
\end{figure}
Since its appearance, KAN has been extensively applied across diverse domains of deep learning \cite{ji2024comprehensive}, including time series forecasting \cite{xu2024kolmogorov}, computer vision and image processing \cite{bodner2024convolutional}, graph neural networks (GNNs) \cite{kiamari2024gkan}, and collaborative filtering. Additionally, it has demonstrated significant strengths in interpretability and symbolic regression \cite{hassan2024bayesian}. Across these research directions, KAN exhibits particularly powerful visualization and interpretability capabilities while inherently mitigating the issue of catastrophic forgetting.

\section{Methods}
\label{sec:method}
Our KCM-collaborate-PLM paradigm is shown in Fig.2. In this chapter, we will detail the structure of this paradigm, which includes: (1) Judge Kan Small Model classifies the input data, and for the data whose confidence $C_{x}$ is greater than the threshold $\epsilon$, we will input it into the large model as a knowledge aid for the small model to the large model; For the data with confidence $C_{x}$ less than $\epsilon$, we input it into the small model, thus reducing the call of the large model. (2) For the small model data of the input big model, we modify it to enhance the efficiency of the big model. (3) For small models, we use knowledge distillation to enhance the accuracy of small models. Below we will elaborate on these three methods.
\begin{figure*}[!h]
  \centering
  \includegraphics[scale=0.4]{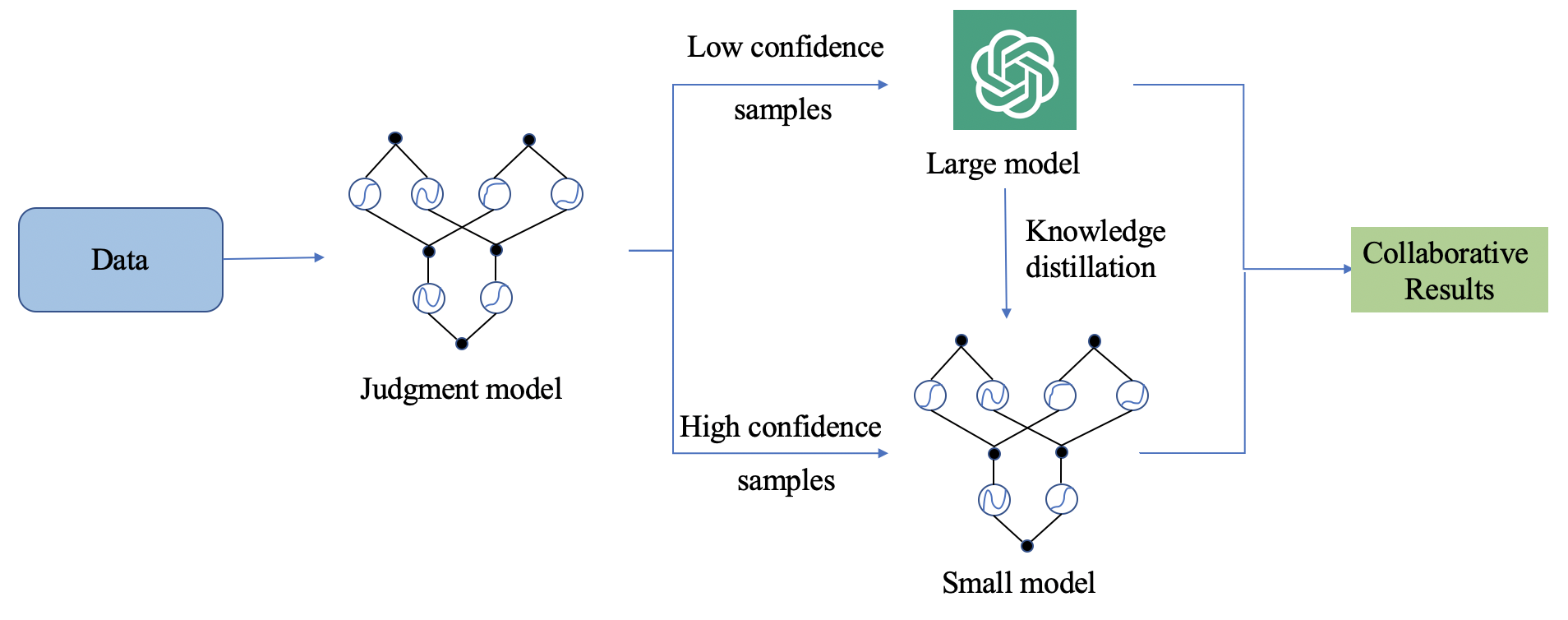}
  \caption{The training process of KCM collaborate PLM. We input all the data into the judgment small model, input the samples with high confidence (representing suitable for small model prediction) into the small model, and input the samples with low confidence (representing suitable for large model prediction) into the large model, and combine the prompt modification of the small model with the knowledge distillation of the large model, and finally combine the prediction results of the two parts to get the result of the collaborative model.}
  \label{fig:fig2}
\end{figure*}
\subsection{KAN-Based Judgment Small Model}
First, for each data x input to judgment model $F_{j}$, the confidence $C_{x}$ is calculated by equation (1). For data with $C_{x}$ less than epsilon, we input it into the large model, and for data with $C_{x}$ greater than epsilon, we input it into the small model.
\begin{equation}
    C_{x} = \frac{{e^{y_i}}}{\Sigma e^{y_n}},y_i\in F_j(x)
\end{equation}
\subsection{Small Model Prompt Modification}
This section mainly discusses how to help the judgment space of large model through the confidence judgment of small model, therefore we propose the method of prompt modification. For the data with low confidence $C_{x}$ that are input into the large model, it is known that these data are not suitable for the classes that the small model is good at, so we can add a label similar to "the confidence of the small model is $C_{x}$" in the prompt, so that the large model pays less attention to these classes that the small model is good at, thus improving efficiency. Then we can calculate the result as follows: 
\begin{equation}
    R_l = F_l(x_i,prompt), y_i\in F_s(x)
\end{equation}
$F_l$ means the large model and $R_l$ is the output result.
\subsection{Large Model Distillation}
This section mainly discusses the influence of large model on small model. Considering the extensive knowledge of large model, the ability boundary of small model can be expanded by knowledge distillation\cite{gou2021knowledge}, so that more samples can be processed. In practice, for the input samples, when the reliability of a specific small model is low and the reliability of a large model is high, the learnable small model will obtain the prediction from the large model, so as to introduce additional knowledge and promote cost reduction. Such a process enables small models that can be learned to handle increasingly diverse samples. On the contrary, small models that can be learned will continue to learn high-confidence samples from specific small models to reduce the hallucination that knowledge from large models may produce.\\
\begin{equation}
    C_{l} = \frac{{e^{y_i}}}{\Sigma e^{y_n}},y_i\in F_l(x)
\end{equation}
Similarly, we calculate the confidence of input data $x$ the large model by Equation 3. For data with confidence greater than $\epsilon$, we distill knowledge by calculating KL-divergence by equation 3 and 4, and iteratively calculate it to update $Fs$. 
\begin{equation}
    L_{ls} = KL(F_s(x), C_l), when C_l>\epsilon
\end{equation}
\begin{equation}
    L_{js} = KL(F_s(x), C_x), whenC_x>\epsilon
\end{equation}
\subsection{Algorithm}
To sum up, we can get the algorithm of KCM training process and inference process, and the whole process of the algorithm is shown separately in Algorithm 1 and Algorithm 2.

\begin{algorithm}[!h]
\SetAlgoLined 
\KwIn{Input data $x$, Large model $F_l$, Judgment model $F_j$}
\KwOut{Small model$F_s$}
$F_s \gets F_j$\;
\For{each sample $x_i \in x$}{
    compute $C_x$ and $C_l$\;
    \If{$C_x > \epsilon$}{
        $x_1 \gets x_i$\; }
    
    \ElseIf{$C_l > \epsilon$}{$x_2 \gets x_i$\;}
    \Else{$x_3 \gets x_i$\; 
    }}
\Indp 
Optimize $F_s$ by KL-divergence\;
\Indm 
\Return $\{F_s\}$
\caption{KCM Training}\label{alg:kcm2}
\end{algorithm}
\begin{algorithm}[!h]
\SetAlgoLined 
\KwIn{Input data $x$, Large model $F_l$, Judgment small model $F_j$, Small model$F_s$}
\KwOut{Prediction result $R$}
\For{each sample $x_i \in x$}{
    compute $C_x$ and $C_l$\;
    \If{$C_x > \epsilon$}{
        $R_i \gets F_j(x_i)$\; }
    
    \ElseIf{$C_l > \epsilon$}{$R_i \gets F_s(x_i)$\;}
    \Else{$R_i \gets F_l(x_i)$\; 
    }}
\Indp 

\Indm 
\Return $\{R\}$
\caption{KCM Inference}\label{alg:kcm2}
\end{algorithm}
\section{Experiments}
\label{sec:exp}
Our experiments run on 2080ti GPU. It should be pointed out that considering the limitation of the experimental environment, the "large model" and "small model" used in the experiment are relative. For example, compared with ChatGPT, BERT is not a large language model, but it is undoubtedly a larger model than KCM as a collaboration, so we also regard BERT as large model in the experiment. We choose hypermeter $\epsilon=0.98$.
\vspace{-1em}
\subsection{Experiments on Language Task}
Firstly, we apply KCM to the language modality, selecting sentiment analysis as the target task and utilizing the Amazon Product Data (APD) \cite{mcauley2015image} as the experimental dataset. APD comprises 20 categories of product reviews annotated with positive or negative sentiment labels. We conduct experiments on 6 categories featuring balanced positive and negative samples, partitioning the dataset into training, validation, and test sets. We employ TKAN \cite{xu2024kolmogorov} as the KAN-based small model, while fine-tuned BERT \cite{devlin2019bert} and ChatGPT serve as backbone large model . Experimental results using BERT and ChatGPT as backbone models are presented in Table 1 and Table 2 respectively.Table 1 demonstrates that the KCM-LLM collaboration framework achieves accuracy improvements across all categories except Kindle, with an average accuracy gain of 2.69($\%$). The Kindle category exhibits a minimal decrease of 0.21($\%$). Furthermore, the large model usage rate averages 81.83($\%$), indicating an approximate 18($\%$) reduction in invocation overhead. This reduction becomes more significant in Table 2, where ChatGPT serves as the backbone model. The KCM-augmented system achieves an average 67.85($\%$) reduction in invocation overhead, substantially decreasing computational resource consumption.\\
Building on this, the KCM-LLM collaboration system using ChatGPT shows slight accuracy decreases of 0.2($\%$) and 0.35($\%$) in the Games and Office categories, respectively, compared to standalone ChatGPT. Meanwhile, it achieves improvements (average about 0.72($\%$)) in the remaining four categories. This indicates that in the language modality, KCM significantly reduces large model invocation frequency—thereby lowering computational resource consumption—while maintaining comparable accuracy. This effect is more significant when employing larger backbone models (e.g. ChatGPT).
\begin{table}[!h]
  \label{tab:table1}
  \caption{BERT+KCM experiment accuracy on APD dataset.}
  \centering
  \begin{tabular}{ccccc}
    \toprule
     Category& BERT($\%$)& KCM($\%$) &LM rate($\%$)\\
    \midrule
    Games & 90.39&95.82 &84.02\\
    Kindle &95.89& 95.68& 73.88 \\
    Baby &92.81&94.45 &83.89 \\
    Movies &90.57 & 93.82 &79.23\\
    Electronics&91.76 & 93.66& 82.12\\
    Office & 90.72&94.83&87.84\\
    \bottomrule
  \end{tabular}
\end{table}
\vspace{-3em}
\begin{table}[!h]
  \label{tab:table2}
  \caption{ChatGPT+KCM experiment accuracy on APD dataset.}
  \centering
  \begin{tabular}{ccccc}
    \toprule
     Category & ChatGPT ($\%$)& KCM($\%$) &LM rate($\%$)\\
    \midrule
    Games & 96.22 &96.02 &33.75\\
    Kindle & 95.65 & 96.68& 34.80 \\
    Baby &96.41 &96.45 &25.66 \\
    Movies &93.42 & 94.88 &28.79\\
    Electronics& 95.41 & 95.76& 36.68\\
    Office & 95.45&95.10&33.24\\
    \bottomrule
  \end{tabular}
\end{table}
\vspace{-2em}
\subsection{Experiments on Visual Task}
In this section, we apply KCM to the visual modality, focusing on long-tail image classification—a challenging, prevalent vision task\cite{zhou2018deep}. Following the setting of \cite{li2022trustworthy}, we partition the CIFAR-100 dataset into head, med, and tail regions based on different numbers of samples. We employ CKAN \cite{bodner2024convolutional} as the KAN-based small model and CLIP \cite{radford2021learning} as the large model, with results summarized in Table 3.Large Model rate(LM rate) mentioned in the table and later refers to the frequency with which the collaborative model invocate the large model. The KCM-assisted collaboration framework achieves: (1) a 3.29$\%$ accuracy improvement over the CKAN small model in the head region; (2) a 1.92$\%$ improvement over the CLIP large model in the medium region; (3) a 6.04$\%$ improvement over CLIP in the tail region. Notably, KCM assistance reduces the LM rate to 59.80$\%$ of the baseline, substantially lowering computational costs.
\vspace{-2em}
\begin{table}[!h]
  \label{tab:table3}
  \caption{Experiment accuracy on CIFAR-100-LT dataset.}
  \centering
  \begin{tabular}{cccc}
    \toprule
    Data&Small($\%$) & Large($\%$)     &  KCM($\%$)   \\
    \midrule
    head & 70.25&60.00 &71.54\\
    med &48.21 & 57.28 & 59.20 \\
    tail&43.28 &57.19 &63.23  \\
    overall &53.25 &58.18 & 64.66 \\
    LM rate&0.00& 100.00 & 59.80\\
    \bottomrule
  \end{tabular}
\end{table}
\vspace{-2em}
\subsection{Experiments on Multimodal Task}
We extend the application of KCM to visual-language multimodal tasks in this section, selecting image captioning as the target task and utilizing MSCOCO dataset\cite{lin2014microsoft} as our experiment dataset. We develop our small model using an encoder-decoder architecture, with CKAN serving as the encoder and TKAN as the decoder. Concurrently, BLIP-2 \cite{li2023blip} as the backbone large model. Experimental results are summarized in Table 4. The results demonstrate that with KCM assistance, the frequency of large model invocation reduces to 62.32$\%$ , while achieving superior performance across BLEU-1 to BLEU-4 benchmarks\cite{papineni2002bleu}. This indicates that the KCM+BLIP-2 collaborative framework not only enhances task performance but also reduces computational cost by 37.68$\%$.
\begin{table}[!h]
  \label{tab:table4}
  \caption{Experiment accuracy on MSCOCO dataset.}
  \centering
  \begin{tabular}{cccc}
    \toprule
     &Small& Large   &  KCM\\
    \midrule
    BLEU-1 & 72.92&73.27 &74.54\\
    BLEU-2 &55.73 & 60.04 & 60.89 \\
    BLEU-3 &41.20 &46.99 &47.22  \\
    BLEU-4 &30.28 &36.11 & 36.42 \\
    LM rate&0.00$\%$& 100.00$\%$ & 62.32$\%$\\
    \bottomrule
  \end{tabular}
\end{table}
\vspace{-1em}
\subsection{KAN-Based Collabroative Model vs. MLP-Based Collabroative Model}
In this section, we demonstrate the advantages of KAN-based collaborative small models (KCM) over MLP-based collaborative small models (MCM) through comparative analysis. Consistent with Section 4.2, we employ long-tail image classification on the CIFAR-100-LT dataset as the application scenario. The MCM architecture follows \cite{chen2024data}, utilizing a ResNet-101 model enhanced with prompt pruning and 2-stage confidence distillation as the collaborative model. Experimental results are summarized in Table 5. The data reveals that KCM achieves higher overall accuracy than MCM in the visual modality, with performance gaps widening toward the tail data. This superiority stems from KAN’s enhanced capability in mitigating catastrophic forgetting, thereby yielding superior accuracy in tail data classification. KAN’s edge-based learnable activations (Fig.1) enable adaptive feature reweighting for tail data, which mitigates catastrophic forgetting by preserving rare patterns during distillation—unlike MLPs’ fixed node activations. Concurrently, KCM reduces large model invocation rate by 6.3 $\%$ compared with MCM. In summary, KCM achieves superior computational efficiency with reduced overhead.

\vspace{-2em}
\begin{table}[!h]
  \label{tab:table5}
  \caption{Ablation Experiment accuracy between MCM and KCM.}
  \centering
  \begin{tabular}{cccc}
    \toprule
    Data& MCM($\%$)   &  KCM($\%$)   \\
    \midrule
    head & 70.99 &71.54\\
    med &57.34  & 59.20 \\
    tail&55.61  &63.23  \\
    LM rate &66.10 &59.80\\
    \bottomrule
  \end{tabular}
\end{table}
\vspace{-2em}

\section{Conclusion}
\label{sec:conclusion}
\vspace{-1em}
With the continuous advancement of pre-trained large models, applications leveraging these models have become increasingly prevalent. However, direct utilization of pre-trained large models via APIs incurs substantial computational costs, constraining their broader adoption. Consequently, researchers have proposed auxiliary small models to collaborate with large models, reducing computational overhead by offloading partial data processing to small models. Our research extends this paradigm: to further harness the complementary strengths of small and large models, we introduce our KAN-Based Collabroative Model (KCM). Compared with MLPs, KANs exhibits marginally greater architectural complexity but delivers superior visualization capabilities, interpretability, and mitigation of catastrophic forgetting. At the scale of large models, the disadvantage of KAN’s higher training costs relative to MLPs is circumvented. We validate our approach across diverse modalities and scenarios, demonstrating that it effectively reduces large model invocation frequency compared to standalone large model methods while enhancing task performance in certain modalities. Furthermore, we confirm that KAN, when employed as a collaborative small model, outperforms MLPs in both computational efficiency and task benchmark performance, demonstrating its superiority as an alternative.
\bibliographystyle{IEEEbib}
\bibliography{refs}

\end{document}